\pdfoutput=1
\documentclass[10pt,journal,compsoc]{IEEEtran}

\usepackage[noadjust]{cite}

\usepackage{multirow}

\usepackage[pdftex]{graphicx}
\usepackage{array}
\usepackage[cmex10]{amsmath}
\usepackage{amssymb}
\usepackage{amsthm}
\usepackage[super]{nth}

\newtheorem{theorem}{Theorem}
\newtheorem{lemma}[theorem]{Lemma}
\newtheorem{innercustomthm}{Theorem}
\newenvironment{customthm}[1]
  {\renewcommand\theinnercustomthm{#1}\innercustomthm}
  {\endinnercustomthm}
\renewenvironment{proof}{{\bfseries Proof.}}{\qed}

\usepackage[caption=false,font=footnotesize,labelfont=sf,textfont=sf]{subfig}

\usepackage{algorithmic}
\usepackage{algorithm}

\newcommand{\SWITCH}[1]{\STATE \textbf{switch} (#1)}
\newcommand{\ENDSWITCH}{\STATE \textbf{end switch}}
\newcommand{\CASE}[1]{\STATE \textbf{case} #1\textbf{:} \begin{ALC@g}}
\newcommand{\ENDCASE}{\end{ALC@g}}

\newcommand{\DEFAULT}{\STATE \textbf{default:} \begin{ALC@g}}
\newcommand{\ENDDEFAULT}{\end{ALC@g}}
\newcommand{\DEFAULTLINE}[1]{\STATE \textbf{default:} }
  
\newcommand{\argmax}{\arg\!\max}
\newcommand{\argmin}{\arg\!\min}

\usepackage{url}
\begin{document}

\title{Ensemble of  Example-Dependent Cost-Sensitive Decision Trees}

\author{Alejandro~Correa~Bahnsen,
        Djamila~Aouada,
        and~Bj\"orn~Ottersten% <-this % stops a space
\IEEEcompsocitemizethanks{\IEEEcompsocthanksitem The authors are with the  Interdisciplinary Centre 
for Security, Reliability and Trust, University of Luxembourg.\protect\\
E-mail: \{alejandro.correa, djamila.aouada, bjorn.ottersten\}@uni.lu}
% \thanks{Manuscript received January 10, 2015; revised February, 2015.}
% \thanks{For informations on obtaining reprints of this article, please send \mbox{e-mail} to:
% 		reprints@ieee.org, and reference the Digital Object Identifier below}
% \thanks{Digital Object Identifier 10.1109\/TKDE.2015}
}

% \markboth{IEEE Transactions on Knowledge and data engineering,~Vol.~, No.~, 2015}
% {Correa Bahnsen \MakeLowercase{\textit{et al.}}: Ensemble of  Example-Dependent Cost-Sensitive 
% Decision Trees}

\IEEEtitleabstractindextext{
\begin{abstract}
	Several real-world classification problems are example-dependent cost-sensi\-tive in nature, 
	where the costs due to misclassification vary between examples and not only within classes. 
	However, standard classification 	methods	do not take these costs into account, and assume a 
	constant cost of misclassification  errors. In previous works, some methods that take into 
  account the financial costs into the training of different algorithms have been proposed, with 
  the example-dependent cost-sensitive decision tree algorithm being the one that gives the 
  highest savings. In this paper we propose a new framework of ensembles of example-dependent 
  cost-sensitive decision-trees. The framework consists in  creating different example-dependent 	
  cost-sensitive decision trees on random subsamples of the training set, and then combining them 
  using three different combination approaches. Moreover, we propose two new cost-sensitive 
  combination approaches; cost-sensitive weighted voting and cost-sensitive stacking, the latter 
  being based on the cost-sensitive logistic   regression   method.
	Finally, using five different databases, from four real-world applications: credit card fraud 	
  detection, churn modeling, credit scoring and direct marketing, we evaluate the 
	proposed method against state-of-the-art example-dependent cost-sensitive techniques, namely, 
  cost-proportionate sampling, Bayes minimum risk and cost-sensitive decision trees.
	The results show that the proposed algorithms have better results for all databases, in the 
  sense of higher savings. 
\end{abstract}

}

\maketitle

% is not selected <OR> if conference mode is selected - because compsoc
% conference papers position the abstract like regular (non-compsoc)
\IEEEdisplaynontitleabstractindextext

% \fi

\IEEEpeerreviewmaketitle

\IEEEraisesectionheading{\section{Introduction}}

\IEEEPARstart{C}{lassification}, in the context of machine learning, deals with the problem of 
	predicting the class of a set of examples given their features. Traditionally, classification 
	methods aim at minimizing the misclassification of examples, in which an example is 
	misclassified if the predicted class is different from the true class. Such a traditional 
	framework assumes that all misclassification errors carry the same cost. This is not the case in 
	many real-world applications. Methods that use different misclassification costs are known as 
	cost-sensitive classifiers. Typical cost-sensitive approaches assume a constant cost for each 
	type of error, in the sense that, the cost depends on the class and is the same among examples 
	\cite{Elkan2001}. 
	
  This class-dependent approach is not realistic in many real-world applications. For 
	example in credit card fraud detection, failing to detect a fraudulent transaction may have an 
  economical impact from a few to thousands of Euros, depending on the particular transaction and 
	card holder \cite{Ngai2011a}. In churn modeling, a model is used for predicting which
	customers are more likely to abandon a service provider. In this context, failing to identify a 
	profitable or unprofitable churner has a significant different economic 
	result~\cite{Verbraken2013}. Similarly, in direct marketing, wrongly predicting that a customer 
	will not accept an offer when in fact he will, may have a different financial impact, as not all 
	customers generate the same profit \cite{Zadrozny2003}. Lastly, in credit scoring, accepting 
	loans from bad customers does not have the same economical loss, since customers have different 
	credit lines, therefore, different profit \cite{Verbraken2014}.
	
	In order to deal with these specific types of cost-sensitive problems, called example-dependent
  cost-sensitive, some methods have been proposed. Standard solutions consist in 
  re-weighting the training examples based on their costs, either by cost-proportionate 
  rejection-sampling \cite{Zadrozny2003}, or cost-proportionate-sampling \cite{Elkan2001}.
  The  rejection-sampling approach consists in selecting a random subset of the 
  training set, by randomly  selecting examples and accepting them with a probability proportional 
  to the misclassification cost of each example. The over-sampling method consists in creating a new 
  training set, by making copies of each example taking into account the misclassification 
  cost. However, cost-proportionate over-sampling increases the training set and it also may 
  result in over-fitting  \cite{Drummond2003}. Also, none of these methods uses take into 
  account the cost of correct classification.
  Moreover, the literature on example-dependent cost-sensitive methods is limited, often because 
  there is a lack of publicly available datasets that fit the problem \cite{MacAodha2013}.
  Recently, we have proposed different methods that take into account the different 
  example-dependent   costs, in particular:  Bayes minimum risk ($BMR$) \cite{CorreaBahnsen2013}, 
  cost-sensitive logistic regression \cite{CorreaBahnsen2014b}, and cost-sensitive decision tree 
  ($CSDT$) \cite{CorreaBahnsen2015}.
  
  The $CSDT$ method is based on a new splitting criteria which is cost-sensitive, used during the 
  tree construction. Then, after the tree is fully grown, it is pruned by using a cost-based 
  pruning criteria. This method was shown to have better results than traditional 
  approaches, in the sense of lower financial costs across different real-world applications, such 
  as in credit card fraud detection and credit scoring. However, the $CSDT$ algorithm only creates 
  one tree in order to make a  classification, and as noted in \cite{Louppe2012}, individual 
  decision trees typically suffer from high variance. A very efficient and simple way to address 
  this flaw is to use them in the context of ensemble methods. 
  
% 	Recently, we proposed a direct cost approach to make the classification decision based on the  
% 	expected costs \cite{CorreaBahnsen2013}. We called this method Bayes minimum risk ($BMR$). It 
% 	consists in quantifying tradeoffs between various decisions using probabilities and the costs  
% 	that accompany such decisions.	Furthermore, in , we proposed a new 
% 	cost-sensitive logistic regression \cite{CorreaBahnsen2014b}. The method consists in introducing 
%   example-dependent costs into a logistic regression, 	by changing the objective function of the 
%   model to one that is 	cost-sensitive.
% 
% 	Furthermore, in \cite{CorreaBahnsen2015} we 
% 	proposed a new example-dependent cost-sensitive decision tree ($CSDT$). The method is based on a 
% 	new splitting criteria which is cost-sensitive, used during the tree construction. Then, after 
% 	the tree is fully grown, it is pruned by using a cost-based pruning criteria.
% 	The CSDT method was shown to have better results than traditional 
% 	approaches, in the sense of lower financial costs across different real-world applications, such 
% 	as in credit card fraud detection and credit scoring.
% 	However, the CSDT algorithm only creates one tree in order to make a 
% 	classification, and as noted in \cite{Louppe2012}, individual decision trees typically suffer 
% 	from high variance. A very efficient and simple way to address this flaw is to use them in the 
% 	context of ensemble methods. 
	  
	Ensemble learning is a widely studied topic in the machine learning community. The main
	idea behind the ensemble methodology is to combine several individual base classifiers in
	order to have a classifier that outperforms each of them \cite{Rokach2009}. Nowadays, 
	ensemble methods are 	one of the most popular and well studied machine learning techniques 
	\cite{Zhou2012}, and it can be noted that since 2009 all the first-place and 
	second-place winners of the KDD-Cup competition\footnote{https\://www.sigkdd.org/kddcup/} used 
	ensemble methods. The core principle in ensemble learning, is to induce random perturbations into 
	the learning procedure in order to produce several different base classifiers from a single 
	training set, then combining the base classifiers in order to make the final prediction.
	In order to induce the random permutations and therefore create the different base classifiers, 
	several methods have been proposed, in particular: bagging \cite{Breiman1996}, 
	pasting~\cite{Breiman1999}, random forests \cite{Breiman2001} and random patches 
	\cite{Louppe2012}. Finally, after  the base 	classifiers are trained, they are typically 
	combined using either 	majority voting, 	weighted 	voting  	or 	stacking~\cite{Zhou2012}.
% 	In the context of cost-sensitive classification, some authors have proposed methods for using 
% 	ensemble techniques, however, all of them focus on class-dependent cost-sensitivity 
% 	\cite{Street2008,Nesbitt2010,Masnadi-shirazi2011,Lomax2013}. Therefore, these methods are not 
% 	well suited for example-dependent cost-sensitive problems. 

  In the context of cost-sensitive classification, some authors have proposed methods for using 
  ensemble techniques. In \cite{Masnadi-shirazi2011}, the authors proposed a framework for 
  cost-sensitive boosting that is expected to minimized the losses by using optimal cost-sensitive 
  decision rules. In \cite{Street2008}, a bagging algorithm with adaptive costs was proposed. In 
  his doctoral thesis, Nesbitt \cite{Nesbitt2010}, proposed a method for cost-sensitive 
  tree-stacking. In this method different decision trees are learned, and then combined in a way 
  that a cost function is minimize. Lastly in \cite{Lomax2013}, a survey of application of 
  cost-sensitive learning with decision trees is shown, in particular including other methods that 
  create cost-sensitive ensembles. However, in all these methods, the  misclassification costs only 
  dependent on the class, therefore, assuming a constant cost across  examples. As a consequence, 
  these methods are not  well suited for example-dependent cost-sensitive  problems. 

	In this paper we propose a new framework of ensembles of example-dependent cost-sensitive 
  decision-trees, by training example-dependent cost-sensitive decision trees using four different 
	random inducer methods and then blending them using three different combination approaches.
	Moreover, we propose two new cost-sensitive combination approaches, cost-sensitive weighted 
	voting and cost-sensitive stacking. The latter being an extension of our previously proposed 
	cost-sensitive logistic regression.	We evaluate the 
	proposed framework using five different databases from four real-world problems. In particular,
	credit card fraud detection, churn modeling, credit scoring and direct marketing. The 
	results show that the proposed method outperforms state-of-the-art 	example-dependent 	
	cost-sensitive methods in three databases, and have a similar result in the other two.	
  Furthermore, our source code, as used for the experiments, is publicly  
	available as part of the \textit{CostSensitiveClassification}\footnote{
	https://github.com/albahnsen/CostSensitiveClassification} library.
    
  The remainder of the paper is organized as follows. In Section 2, we explain the background 
	behind example-dependent cost-sensitive classification and ensemble learning. In Section 3, we 
	present the proposed ensembles of cost-sensitive decision-trees framework. Moreover, in 
	\mbox{Section 4}, we prove theoretically that combining individual cost-sensitive classifiers 
	achives better 	results in the sense of higher financial savings. Then the experimental setup 
	and the different 	datasets are described in Section 5.  Subsequently, the proposed algorithms 
	are evaluated and compared against state-of-the-art methods on these	different datasets. 
	Finally, conclusions are given in 	\mbox{Section 7}.
	
\section{Background and problem formulation}

This work is related to two groups of research in the field of machine learning: (i) 
example-dependent cost-sensitive classification, and (ii) ensemble learning.

\subsection{Example-dependent cost-sensitive classification} \label{sec:edcs}
	Classification deals with the problem	of predicting the class $y_i$ of a set $\mathcal{S}$ of 
	examples or instances, given their $k$ features \mbox{$\mathbf{x}_i \in \mathbb{R}^k$}. 
	The objective is to construct a function $f(\cdot)$ that makes a prediction $c_i$ of the 
	class of each example using its variables $\mathbf{x}_i$. 	Traditionally, machine 
	learning classification	methods are designed to minimize some sort 	of misclassification measure 
	such as the F1Score \cite{Hastie2009};	therefore, assuming that different misclassification 
	errors have the same cost. As discussed before, this is not suitable in many real-world 
	applications. Indeed, two classifiers with equal misclassification rate but different numbers of 
	false positives and false negatives do 	not have the same impact on cost since \mbox{$C_{FP_i} 
	\ne C_{FN_i}$};	therefore, there is a need 	for a measure that takes into account 	the actual 
	costs of each example $i$.
		
	In this context, binary classification costs can be represented using a 2x2 cost matrix 
	\cite{Elkan2001}, that introduces the costs associated with two types of correct classification, 
	true positives ($C_{TP_i}$), true negatives ($C_{TN_i}$), and the two types of misclassification 
	errors, false positives ($C_{FP_i}$), false negatives ($C_{FN_i}$),	as defined in 
	\tablename{ \ref{t:costmat}}.
  Conceptually, the cost of correct classification should always be lower than the cost of 
	misclassification. These are referred to as the \textit{reasonableness conditions} 
	\cite{Elkan2001}, 	and are defined as $C_{FP_i} > C_{TN_i}$ and $C_{FN_i} > C_{TP_i}$. 
	
	\begin{table}[b]
		\caption{Classification cost matrix}
		\centering
		\begin{tabular}{c|c|c}
			\multicolumn{1}{c|}{}  & Actual Positive& Actual Negative \\
			\multicolumn{1}{c|}{} & $y_i=1$& $y_i=0$ \\
			\hline
			Predicted Positive 		& \multirow{ 2}{*}{$C_{TP_i}$} & \multirow{ 2}{*}{$C_{FP_i}$} \\
			$c_i=1$ & &\\
			\hline
			Predicted Negative  	& \multirow{ 2}{*}{$C_{FN_i}$} & \multirow{ 2}{*}{$C_{TN_i}$} \\
			$c_i=0$ & &\\
		\end{tabular}\label{t:costmat}
	\end{table}  

	Let $\mathcal{S}$ be a set of $N$ examples $\mathbf{x}_i$, where 
	each example is represented by the augmented feature vector \mbox{$\mathbf{x}_i^*=[\mathbf{x}_i, 
	C_{TP_i}, C_{FP_i}, C_{FN_i}, C_{TN_i}]$} and labelled using the class label $y_i$. A classifier 
	$f$ 	which generates the 	predicted label $c_i$ for each example $i$ 	is trained using the set 
	$\mathcal{S}$.  Using the cost matrix, an example-dependent cost statistic 
  \cite{CorreaBahnsen2013}, is defined as:
	\begin{align}\label{eq:cost}
		Cost(f(\mathbf{x}_i^*)) =& y_i(c_i C_{TP_i} + (1-c_i)C_{FN_i}) + \nonumber \\  
		& (1-y_i)(c_i C_{FP_i} + (1-c_i)C_{TN_i}) ,
	\end{align}
	leading to a total cost of:
	\begin{equation}\label{eq:cost_total}
	 	 Cost(f(\mathcal{S})) = \sum_{i=1}^N Cost(f(\mathbf{x}_i^*)).
	\end{equation}

	\noindent However, the total cost may not be easy to interpret. In~\cite{Whitrow2008}, a 
	\textit{normalized} cost measure was proposed, by dividing the total cost by the theoretical 
	maximum cost, which is the cost of misclassifying every example. The \textit{normalized} cost is 
	calculated using
  \begin{align}\label{eq:ncost}
    Cost_n(f(\mathcal{S})) = \frac{Cost(f(\mathcal{S}))}
    {\sum_{i=1}^N C_{FN_i} \cdot \mathbf{1}_0(y_i) 
    +  C_{FP_i} \cdot \mathbf{1}_1(y_i)  },
  \end{align} 
  where $\mathbf{1}_c(z)$ is an indicator function that takes the value of one if $z = c$ and 
	zero if $z \ne c$.
  
	We proposed a similar approach in \cite{CorreaBahnsen2014b}, where the savings corresponding to 
	using an algorithm  are defined as the cost of the algorithm versus the cost of using no 
	algorithm at all. To do that, the cost of the costless class is defined as % TODO: Change costless
	\begin{equation}
		Cost_l(\mathcal{S}) = \min \{Cost(f_0(\mathcal{S})), Cost(f_1(\mathcal{S}))\},
	\end{equation}
	where 
	\begin{equation}\label{eq:f_a}
		f_a(\mathcal{S}) = \mathbf{a}, \text{ with } a\in \{0,1\}.
	\end{equation}

	\noindent The cost improvement can be expressed as the cost of savings as compared with 
  $Cost_l(\mathcal{S})$. 
  \begin{equation}\label{eq:savings}
    Savings(f(\mathcal{S})) = \frac{ Cost_l(\mathcal{S}) - Cost(f(\mathcal{S}))}
		{Cost_l(\mathcal{S})}.
  \end{equation}

\subsection{Ensemble learning}

	Ensemble learning is a widely studied topic in the machine learning community. The main idea 
	behind the ensemble methodology is to combine several individual classifiers, referred to as base 
	classifiers, in order to have a classifier that outperforms everyone of them \cite{Rokach2009}.
	There are three main reasons regarding why ensemble methods perform better than 
	single models: statistical, computational and representational \cite{Dietterich2000a}.
	First, from a statistical point of view, when the learning set is too small, an algorithm can 
	find several good models within the search space, that arise to the same performance on the 
	training set $\mathcal{S}$. Nevertheless, without a validation set, there is risk of choosing 
	the wrong model. The second reason is computational; in general, algorithms rely on some local 
	search optimization and may get stuck in a local optima. Then, an ensemble may solve this by 
	focusing different algorithms to different spaces across the training set.
	The last reason is representational. In most cases, for a learning set of finite size, the 
	true function $f$ cannot be represented by any of the candidate models. By combining several 
	models in an ensemble, it may be possible to obtain a model with a larger coverage across the 
	space of representable functions.

	The most typical form of an ensemble is made by combining $T$ different base classifiers.
	Each  base classifier $M(\mathcal{S}_j)$ is trained by applying algorithm $M$ to a random subset 
	$\mathcal{S}_j$ of the training set $\mathcal{S}$.  %  $\mathcal{S_j} = RI(\mathcal{S})$
	For simplicity we define $M_j \equiv 	M(\mathcal{S}_j)$ for $j=1,\dots,T$, and 
	$\mathcal{M}=\{M_j\}_{j=1}^{T}$ a set of base classifiers.
	Then, these models are combined using majority voting to create the ensemble $H$ as follows
	\begin{align}\label{eqn:majority-vote}
		f_{mv}(\mathcal{S},\mathcal{M}) = \argmax_{c \in \{0,1\}} \sum_{j=1}^T 
		\mathbf{1}_c(M_j(\mathcal{S})).
	\end{align}
	% http://en.wikipedia.org/wiki/Indicator_function
	
	\noindent Moreover, if we assume that each one of the $T$ base classifiers has a probability 
  $\rho$ of being correct, the probability of an ensemble making the correct decision, denoted by 
  $P_c$, can be calculated using the binomial \mbox{distribution \cite{Hansen1990}}
	\begin{equation}\label{eq:prob}
		P_c = \sum_{j>T/2}^{T} {{T}\choose{j}} \rho^j(1-\rho)^{T-j}.
	\end{equation}
	Furthermore, as shown in \cite{Lam1997}, if $T\ge3$ then:
% 	\begin{itemize}
% 		\item If $\rho>0.5$, then $ \lim_{T \to  \infty} P_c=1 $
% 		\item If $\rho<0.5$, then $ \lim_{T \to  \infty} P_c=0 $ 
% 		\item If $\rho=0.5$, then $P_c=0.5$ for any $T$,
% 	\end{itemize}
	\begin{equation}\label{eq:Pc}
  \lim_{T \to  \infty} P_c= \begin{cases} 
            1  &\mbox{if } \rho>0.5 \\ 
            0  &\mbox{if } \rho<0.5 \\ 
            0.5  &\mbox{if } \rho=0.5 ,
            \end{cases}
  \end{equation}
	leading to the conclusion that 
	\begin{equation}\label{eq:Pc2}
  \rho \ge 0.5 \quad \text{and} \quad T\ge3 \quad \Rightarrow \quad P_c\ge \rho.
  \end{equation}

% 	\begin{figure}[!t]
% 		\centering
% 		\includegraphics[width=3in]{fig_motivation_ensemble}
% 		\label{fig_first_case}
% 		\caption{Fundamental reasons why an ensemble may work better than a single classifier 
% 		\cite{Dietterich2000a}.}
% 		\label{fig:motivation_ensemble}
% 	\end{figure}
	
\section{Ensembles of cost-sensitive decision-trees}

	In this section, we present our proposed framework for ensembles of example-dependent 
	cost-sensitive decision-trees ($ECSDT$). The framework is based on expanding our previous 
	contribution on example-dependent cost-sensitive decision trees ($CSDT$)
	\cite{CorreaBahnsen2015}. In particular, we create many different $CSDT$ on random subsamples 
	of the training set, and then combine them using different combination methods. Moreover, we 
	propose two new cost-sensitive combination approaches, cost-sensitive weighted voting and 
	cost-sensitive stacking. The latter being an extension of our previously proposed cost-sensitive 
	logistic regression~($CSLR$)~\cite{CorreaBahnsen2014b}.
	
	The remainder of the section is organized as follows: First, we introduce the example-dependent 
	cost-sensitive decision tree. Then we present the different random inducers and combination 
	methods. Finally, we define our proposed algorithms.

\subsection{Cost-sensitive decision tree  ($CSDT$)}
	
	Introducing the cost into the training of a decision tree has been a widely study way of making 
  classifiers cost-sensitive \cite{Lomax2013}. However, in most cases, approaches that have been 
  proposed only deal with the problem when the cost depends on the class and not on the example
	\cite{Draper1994,Ting2002,Ling2004,Li2005,Kretowski2006,Vadera2010}. 
	In \cite{CorreaBahnsen2015}, we proposed an example-dependent cost-sensitive decision trees 
  ($CSDT$) algorithm, that takes into account the example-dependent costs during the training and 
  pruning of a tree.
	
	In the $CSDT$ method, a new splitting criteria is 
	used during the tree construction. In particular, instead of using a traditional splitting 
	criteria such as Gini, entropy or misclassification, the cost as defined in (\ref{eq:cost}), of 
	each tree node is calculated, and the gain of using each split evaluated as the decrease in 
	total cost of the algorithm.

	The cost-based impurity measure is defined by comparing the costs when all the 
	examples in a leaf are classified as negative and as positive,
	\begin{equation}\label{eq:cost_impurity}
		I_c(\mathcal{S}) = \min \bigg\{ Cost(f_0(\mathcal{S})), Cost(f_1(\mathcal{S})) \bigg\}.
	\end{equation}
	Then, using the cost-based impurity, the gain of using 	the splitting rule $(\mathbf{x}^j,l^j)$, 
	that is the rule of splitting the set $\mathcal{S}$ on feature $\mathbf{x}^j$ on value $l^j$,  
	is calculated as: 
	\begin{equation}\label{eq:gain}
		 Gain(\mathbf{x}^j,l^j)=I_c(\mathcal{S})-
		 \frac{\vert \mathcal{S}^l \vert}{\vert \mathcal{S} \vert}I_c(\mathcal{S}^l)
		 -\frac{\vert \mathcal{S}^r \vert}{\vert \mathcal{S} \vert}I_c(\mathcal{S}^r),
	\end{equation} 
	where 
	$\mathcal{S}^l = \{\mathbf{x}_i^* \vert \mathbf{x}_i^* \in \mathcal{S} \wedge x^j_i \le l^j \}$, 
		$\mathcal{S}^r = \{\mathbf{x}_i^* \vert \mathbf{x}_i^* \in \mathcal{S} \wedge x^j_i > l^j \}$,
	and $\vert \cdot \vert$ denotes the cardinality. Afterwards, using the cost-based gain measure, a 
  decision tree is grown until no further splits can be made.

	Lastly, after the tree is constructed, it is pruned by using a cost-based pruning criteria 
	\begin{equation}\label{eq:cost_pruning}
		PC_{c} =  Cost(f(S)) - Cost(f^*(S)),
	\end{equation}
	where $f^*$ is the classifier of the tree without the pruned node.

\subsection{Algorithms}

	With the objective of creating an ensemble of example-dependent cost-sensitive decision 
	trees, we first create $T$ different random subsamples $\mathcal{S}_j$ for $j=1,\dots,T$, of the 
	training 	set $\mathcal{S}$, and train a $CSDT$ algorithm on each one. In particular we create 
	the different subsets using four different methods: bagging \cite{Breiman1996}, pasting 
	\cite{Breiman1999}, random forests \cite{Breiman2001} and random patches \cite{Louppe2012}. 
	
	In bagging \cite{Breiman1996}, base classifiers are built on randomly drawn bootstrap subsets of 
  the original data, hence producing different base classifiers. Similarly, in pasting 
  \cite{Breiman1999}, the base classifiers are built on random 	samples without replacement from 
  the training set. In random forests \cite{Breiman2001}, using decision trees as the base learner, 
  bagging 	is extended and 	combined 	with a 	randomization of the input features that  are used 
  when 	considering candidates 	to split 		internal nodes. In particular, instead of looking for  
  the best 	split among all 	features, the 	algorithm selects, at each node, a random subset of 
  features 	and then determines 	the best split only over 	these features. In the random patches 	
  algorithm \cite{Louppe2012}, base classifiers are created by randomly 		drawn bootstrap subsets 
  of both examples and features.
	
	Lastly, the base classifiers are combined using either majority voting, cost-sensitive weighted 
	voting and cost-sensitive stacking. Majority voting consists in collecting the predictions of 
	each base classifier and selecting the decision with the highest number of votes, see 
	(\ref{eqn:majority-vote}).

	\subsubsection*{Cost-sensitive weighted voting}

	This method is an extension of weighted voting. First, in the traditional approach, a 
	similar comparison of the votes of the base classifiers is made, but giving a weight $\alpha_j$ 
	to each classifier $M_j$ during the voting phase \cite{Zhou2012}
	\begin{align} \label{eqn:weighted-majority-vote}
		f_{wv}(\mathcal{S},\mathcal{M}, \mathbf{\alpha})
		=\argmax_{c \in \{0,1\}} \sum_{j=1}^T \alpha_j \mathbf{1}_c(M_j(\mathcal{S})),
	\end{align}
	where $\mathbf{\alpha}=\{\alpha_j\}_{j=1}^T$.
	The calculation of $\alpha_j$ is related to the performance of each classifier $M_j$.
	It is usually defined as the normalized misclassification error 	$\epsilon$ of the base 
	classifier $M_j$ 	in the out of bag set 	$\mathcal{S}_j^{oob}=\mathcal{S}-\mathcal{S}_j$
	\begin{equation}
		\alpha_j=\frac{1-\epsilon(M_j(\mathcal{S}_j^{oob}))}{\sum_{j_1=1}^T 
		1-\epsilon(M_{j_1}(\mathcal{S}_{j_1}^{oob}))}.
	\end{equation}
	However, as discussed in Section \ref{sec:edcs}, the misclassification measure is not suitable in 
	many real-world classification problems. We herein propose a method to calculate the weights 
$\alpha_j$ 
  taking into account the actual savings of the classifiers. Therefore using (\ref{eq:savings}), we 
  define
	\begin{equation}
		\alpha_j=\frac{Savings(M_j(\mathcal{S}_j^{oob}))}
		{\sum_{j_1=1}^T Savings(M_{j_1}(\mathcal{S}_j^{oob}))}.
	\end{equation}
	This method guaranties that the base classifiers that contribute to a higher increase in savings 
	have more importance in the ensemble.
	
	\subsubsection*{Cost-sensitive stacking}
	
	The staking method consists in combining the different base classifiers by learning a 
	second level algorithm on top of them \cite{Wolpert1992}. In this framework, once the base 
	classifiers are constructed using the training set 	$\mathcal{S}$, a new set is constructed 
	where the output of the base classifiers 	are now considered as the features while keeping the 
	class labels.
	
	Even though there is no restriction on which algorithm can be used as a second level learner, 
	it is common to use a linear model \cite{Zhou2012}, such as 
	\begin{align}\label{eqn:stacking}
		f_s(\mathcal{S},\mathcal{M},\mathcal{\beta}) =
		g \left( \sum_{j=1}^T \beta_j M_j(\mathcal{S}) \right),
	\end{align}
	where $\mathcal{\beta}=\{\beta_j\}_{j=1}^T$, and $g(\cdot)$ is the sign function 
	\mbox{$g(z)=sign(z)$} in the case of a linear regression or the sigmoid function, defined 
	as \mbox{$g(z)=1/(1+e^{-z})$}, in the case of a logistic regression. 
	
	Moreover, following the logic used in \cite{Nesbitt2010}, we propose learning the set of  
	parameters $\mathcal{\beta}$  using our proposed cost-sensitive logistic regression ($CSLR$) 
	\cite{CorreaBahnsen2014b}. The $CSLR$ algorithm consists in introducing example-dependent costs 
	into a logistic regression, by changing the objective function of the model to one that is 
	cost-sensitive. For the specific case of cost-sensitive stacking, we define the cost function as: 
  \begin{align}\label{eq:CSLR}
		&J(\mathcal{S},\mathcal{M},\beta)= \nonumber \\
		& \sum_{i=1}^{N} \bigg[ y_i\bigg( 
		f_s(\mathbf{x}_i,\mathcal{M},\mathcal{\beta}) \cdot(C_{TP_i} - C_{FN_i}) + C_{FN_i} \bigg) + 
		\nonumber \\
    & (1-y_i)\bigg( f_s(\mathbf{x}_i,\mathcal{M},\mathcal{\beta}) \cdot(C_{FP_i} - C_{TN_i}) 
			+C_{TN_i} \bigg) \bigg].
  \end{align}
  Then, the parameters $\beta$ that minimize the logistic cost function are used in order to 
	combine the different base classifiers. However, as discussed in \cite{CorreaBahnsen2014b}, 
	this cost function is not convex for all possible cost matrices, therefore, we use genetic 
	algorithms to minimize it.
	
	Similarly to cost-sensitive weighting, this method guarantees that the base classifiers that 
	contribute to a higher increase in savings have more importance in the ensemble. Furthermore, 
	by learning an additional second level cost-sensitive method, the combination is made such that 
	the overall 	savings measure is maximized.
	
	Finally, Algorithm 1 summarizes the proposed $ECSDT$ methods. In total, we evaluate 12 different 
	algorithms, as four different random inducers (bagging, pasting, random forest and random 
	patches) and three different combinators (majority voting, cost-sensitive weighted voting and 
	cost-sensitive stacking) can be selected in order to construct the cost-sensitive ensemble.

\begin{algorithm}[!t]
\caption{The proposed $ECSDT$ algorithms.} 
\label{algo:bagging}
\begin{algorithmic}  
\REQUIRE $CSDT$ (an example-dependent cost-sensitive decision tree algorithm), $T$ the number of 
iterations, $\mathcal{S}$ the training set, $inducer$, $N_e$ number of 
examples for each base classifier, $N_f$ number of examples for each base classifier, 
$combinator$.
\STATE \textbf{Step 1:} Create the set of base classifiers
\FOR{$j  \leftarrow 1$ to $T$}
	\SWITCH {$inducer$}
	\CASE {Bagging}
		\STATE $\mathcal{S}_j \leftarrow $ Sample $N_e$ examples from $\mathcal{S}$ with replacement.
	\ENDCASE
	\CASE {Pasting}
		\STATE $\mathcal{S}_j \leftarrow $ Sample $N_e$ examples from $\mathcal{S}$ without replacement.
	\ENDCASE
	\CASE {Random forests}
		\STATE $\mathcal{S}_j \leftarrow $ Sample $N_e$ examples from $\mathcal{S}$ with replacement.
	\ENDCASE
	\CASE {Random patches}
		\STATE $\mathcal{S}_j \leftarrow $ Sample $N_e$ examples and $N_f$ features from $\mathcal{S}$ 
with replacement.
	\ENDCASE
\ENDSWITCH
	\STATE $M_j \leftarrow CSDT(\mathcal{S}_j)$
	\STATE $\mathcal{S}_j^{oob} \leftarrow \mathcal{S}-\mathcal{S}_j$
	\STATE $\alpha_j \leftarrow Savings( M_j(\mathcal{S}_j^{oob}))$
\ENDFOR

\STATE \textbf{Step 2:} Combine the different base classifiers
	\SWITCH {$combinator$}
	\CASE {Majority voting}
		\STATE $H \leftarrow f_{mv}(\mathcal{S}, \mathcal{M})$
	\ENDCASE
	\CASE {Cost-sensitive weighted voting}
		\STATE $H \leftarrow f_{wv}(\mathcal{S}, \mathcal{M}, \alpha)$
	\ENDCASE
	\CASE {Cost-sensitive stacking}
		\STATE $\mathcal{ \beta } \leftarrow \argmin_{\beta \in \mathbb{R}^T} 
		J(\mathcal{S},\mathcal{M},\beta)$
		\STATE $H \leftarrow f_{s}(\mathcal{S}, \mathcal{M},\mathcal{\beta})$
	\ENDCASE
	\ENSURE $H$ (Ensemble of cost-sensitive decision trees)
\end{algorithmic}  
\end{algorithm}

\section{Theoretical analysis of the cost-sensitive ensemble}

	Although the above proposed algorithm is simple, there is %little 
	no work that has formally 
	investigated ensemble performance in terms other than accuracy. In this section, our aim is to 
	prove theoretically that combining individual cost-sensitive classifiers achieves better results 
	in the sense of higher savings.
	
	We denote $\mathcal{S}_a$, where $a\in \{0,1\}$, as the subset of $\mathcal{S}$ 
	where the examples belong to the class $a$:
	\begin{equation}\label{eq:S_a}
		\mathcal{S}_a = \{\mathbf{x}_i^* \vert y_i = a, i \in 1,\dots,N\},
	\end{equation}
	where $\mathcal{S}=\mathcal{S}_0 \cup \mathcal{S}_1$, $\mathcal{S}_0 \cap \mathcal{S}_1 = 
	\varnothing$, and $N_a=\vert \mathcal{S}_a \vert$. Also, we define the average cost of the base 
	classifiers as:
	\begin{align}\label{eq:avg_cost}
		\overline{Cost} (\mathcal{M}(\mathcal{S}))= \frac{1}{T} \sum_{j=1}^{T} Cost(M_j(\mathcal{S})). 
	\end{align}
	
	\noindent Firstly, we prove the following lemma that states the cost of an ensemble $H$ on the 
	subset $\mathcal{S}_a$ is lower than the average cost of the base classifiers on the same set for 
	$a 	\in \{0,1\}$.

	\begin{lemma}\label{lemma1}
	Let $H$ be an ensemble of $T\ge3$ classifiers $\mathcal{M}=\{M_1, M_2,\dots,M_T\}$, and 
	$\mathcal{S}$ a testing set of size $N$. If each one of the base classifiers has a probability 
	of being correct higher or equal than one half, $\rho \ge \frac{1}{2}$, and the 
	\textit{reasonableness conditions} of the cost matrix are satisfied, then the following holds true
	\begin{align}\label{eq:lemma}
		Cost(H(\mathcal{S}_a)) &\le \overline{Cost} (\mathcal{M}(\mathcal{S}_a)) , \quad a  \in\{0,1\}, 
	\end{align}
	\end{lemma}
 
	\noindent\begin{proof}
	First, we decompose the total cost of the ensemble by applying equations (\ref{eq:cost}) 
	and (\ref{eq:cost_total}). Additionally, we separate the analysis for $a=0$ and $a=1$:

	\textbullet\ $a=0:$
	\begin{align}
	Cost(H(\mathcal{S}_0)) =& \sum_{i=1}^{N_0} y_i(c_i C_{TP_i} + (1-c_i)C_{FN_i})+ \nonumber \\  
		& (1-y_i)(c_i C_{FP_i} + (1-c_i)C_{TN_i}) . 
	\end{align}
	
	\noindent Moreover, we know from (\ref{eq:prob}) that the probability of an ensemble 
	making the right decision, i.e., $y_i=c_i$, for any given 	example, is equal to $P_c$. 
	Therefore, we can use this probability to estimate the expected savings of an ensemble: 
	\begin{align}\label{eq:21}
		Cost(H(\mathcal{S}_0)) &= \sum_{i=1}^{N_0} P_c C_{TN_i} +(1-P_c)C_{FP_i}.
	\end{align}
	
	\textbullet\ $a=1:$
	
	\noindent In the case of $\mathcal{S}_1$, and following the same logic as when $a=0$, the cost of 
	an 	ensemble is:
	\begin{align}\label{eq:22}
		Cost(H(\mathcal{S}_1)) &= \sum_{i=1}^{N_1} P_c C_{TP_i} + (1-P_c)C_{FN_i}.  
	\end{align}

	\noindent The second part of the proof consists in analyzing the right hand side of 
	(\ref{eq:lemma}), 	specifically, the average cost of the base classifiers on set 
	$\mathcal{S}_a$. To do that, with the help of (\ref{eq:cost_total}) and (\ref{eq:avg_cost}), we 
	may express the	average cost of the base classifiers as:
	\begin{align}
		\overline{Cost} (\mathcal{M}(\mathcal{S}_a)) = \frac{1}{T} \sum_{j=1}^{T} \sum_{i=1}^{N_a} 
		Cost(M_j(\mathbf{x}_i^*)).  
	\end{align}	
	
	\noindent We define the set of base classifiers that make a negative prediction as
	\mbox{$\mathcal{T}_{i0}=\{M_j(\mathbf{x}_i^*) \vert M_j(\mathbf{x}_i^*) = 0, j \in 1,\dots,T\}$},
	similarly, the set of classifiers that make a positive prediction as
	\mbox{$\mathcal{T}_{i1}=\{M_j(\mathbf{x}_i^*) \vert M_j(\mathbf{x}_i^*) = 1, j \in 1,\dots,T\}$}.
	Then, by taking the cost of negative and positive predictions from (\ref{eq:f_a}), the 
	average cost of the base learners becomes:
	\begin{align}
		\overline{Cost} (\mathcal{M}(\mathcal{S}_a)) =& 
		\frac{1}{T} \sum_{i=1}^{N_a} \bigg( \vert \mathcal{T}_{i0} \vert \cdot 
		Cost(f_0(\mathbf{x}_i^*)) \nonumber \\
		& + \vert \mathcal{T}_{i1} \vert \cdot Cost(f_1(\mathbf{x}_i^*)) \bigg).
	\end{align}	

	\newpage
	\noindent We separate the analysis for $a=0$ and $a=1$:
	
	\textbullet\ $a=0:$
	
	\begin{align}
		\overline{Cost} (\mathcal{M}(\mathcal{S}_0)) =& \sum_{i=1}^{N_0} \bigg( 
		\frac{\vert \mathcal{T}_{i0} \vert}{T} \cdot C_{TN_i}
		+ \frac{\vert \mathcal{T}_{i1} \vert}{T} \cdot C_{FP_i}\bigg).
	\end{align}
	Furthermore, we know from (\ref{eq:prob}) that an average base classifier will have a correct 
	classification probability of $\rho$, then $\frac{\vert \mathcal{T}_{i0} \vert}{T}=\rho$, leading 
	to:
	\begin{align}\label{eq:26}
		\overline{Cost} (\mathcal{M}(\mathcal{S}_0)) =& \sum_{i=1}^{N_0}  
		\rho \cdot C_{TN_i} + (1-\rho) \cdot C_{FP_i} .
	\end{align}

	\textbullet\ $a=1:$
 
 \noindent Similarly, for the set $\mathcal{S}_1$, the average classifier will have a correct 
	classification probability of $\rho$, then $\frac{\vert \mathcal{T}_{i1} \vert}{T}=\rho$. 
	
	\noindent Therefore,
	\begin{align}\label{eq:27}
		\overline{Cost} (\mathcal{M}(\mathcal{S}_1)) =& \sum_{i=1}^{N_1}  
		\rho \cdot C_{TP_i} + (1-\rho) \cdot C_{FN_i} .
	\end{align}
 
	\noindent Finally, by replacing in (\ref{eq:lemma}) the expected savings of an  ensemble with 
  (\ref{eq:21}) for $a=0$ and (\ref{eq:22}) for $a=1$, and the average cost of the base learners 
  with  (\ref{eq:26}) for $a=0$ and (\ref{eq:27}) for $a=1$, (\ref{eq:lemma}) is rewritten as:
	
	\noindent for $a=0$:
  \begin{align}\label{eq:lemma1}
		\sum_{i=1}^{N_0} P_c C_{TN_i} +(1-P_c)C_{FP_i} \le 
		\sum_{i=1}^{N_0} \rho C_{TN_i} + (1-\rho)  C_{FP_i},
	\end{align}
	for $a=1$:
  \begin{align}\label{eq:lemma2}
		\sum_{i=1}^{N_1} P_c C_{TP_i} + (1-P_c)C_{FN_i} \le 
		\sum_{i=1}^{N_1}  \rho C_{TP_i} + (1-\rho)  C_{FN_i}.
	\end{align}
 
  \noindent Since $\rho \ge \frac{1}{2}$ , then $P_c\ge\rho$ from (\ref{eq:Pc2}), and using the 
  \textit{reasonableness conditions} described in Section \ref{sec:edcs}, i.e, $C_{FP_i} > 
  C_{TN_i}$ and   $C_{FN_i} > C_{TP_i}$, we find that (\ref{eq:lemma1}) and (\ref{eq:lemma2}) are 
  True.
\end{proof}

	Lemma \ref{lemma1} separates the costs on sets $\mathcal{S}_0$ and 	$\mathcal{S}_1$. We are 
	interested in analyzing the overall savings of an ensemble. In this direction, we demonstrate 
	in the following theorem, that the expected savings of an ensemble of classifiers are higher 
	than the expected average savings of the base learners.
	
	\begin{customthm}{1}\label{theorem1}
	Let $H$ be an ensemble of $T\ge3$ classifiers $\mathcal{M}=\{M_1,\dots,M_T\}$, and $\mathcal{S}$ 
  a testing set of size $ N $, then the expected savings of using $H$ in 
	$\mathcal{S}$ are lower than the average expected savings of the base classifiers, in other words,
	\begin{equation}\label{eq:theorem}
		Savings(H(\mathcal{S})) \ge \overline{Savings}(\mathcal{M}(\mathcal{S})). 
	\end{equation}
	\end{customthm}

	\begin{proof}
	Given (\ref{eq:savings}), (\ref{eq:theorem}) is equivalent to
	\begin{equation}\label{eq:theorem2}
		Cost(H(\mathcal{S})) \le \overline{Cost} (\mathcal{M}(\mathcal{S})). 
	\end{equation}

	\noindent Afterwards, by applying the cost definition (\ref{eq:cost}), and grouping the 
	sets of negative and positive examples using (\ref{eq:S_a}), (\ref{eq:theorem2}) becomes
	\begin{equation}
		\sum_{a\in \{0,1\}} Cost(H(\mathcal{S}_a)) \le \sum_{a\in \{0,1\}} \overline{Cost} (\mathcal{M} 
		(\mathcal{S}_a)),
	\end{equation}
	
	\noindent which can be easily proved using Lemma \ref{lemma1}, since, if the cost of an ensemble 
	$H$ is 	lower than the average cost of the base classifiers on both $\mathcal{S}_0$ and 
	$\mathcal{S}_1$, 	implies that it is also lower on the sum of the cost on both sets, 
	therefore, proving Theorem~\ref{theorem1}.
	\end{proof}

\section{Experimental setup}

	In this section we present the datasets used to evaluate the propose Ensembles of 
	Example-Dependent Cost-Sensitive Decision-Trees algorithms. We used five datasets from four 
	different real world example-dependent cost-sensitive problems: Credit card fraud detection, churn 
	modeling, credit scoring and direct marketing.

	For each dataset we used a pre-defined a cost matrix that we previously proposed in different 
	publications.	Additionally, we perform an under-sampling, cost-proportionate rejection-sampling 
	and cost-proportionate over-sampling procedures. 

\subsection{Credit card fraud detection}
	
	A credit card fraud detection algorithm, consist in identifying those transactions with a 
	high probability of being fraud, based on historical fraud patterns. Different detection 
	systems that are based on machine learning techniques have been successfully used for this 
	problem, for a review see \cite{Ngai2011a}.

	Credit card fraud detection is by definition a cost sensitive problem, since the cost of failing 
	to detect a fraud is significantly different from the one when a false alert is made. We used 
	the fraud detection example-dependent cost matrix we proposed in \cite{CorreaBahnsen2013},	in 
	which the cost of failing to detect a fraud is equal to the amount of the transaction ($Amt_i$),	
	and the costs of correct classification and false positives is equal to the administrative cost 
	of	investigating a fraud alert ($C_a$). The cost table is presented in \tablename{ 
	\ref{tab:f_mat}}. For a further discussion see \cite{CorreaBahnsen2013,CorreaBahnsen2014}. 

	For this paper we used a dataset provided by a large European card processing company. The 
	dataset consists of fraudulent and legitimate transactions made with credit and debit cards 
	between January 2012 and June 2013. The total dataset contains 236,735 individual transactions, 
	each one with 27 attributes, including a fraud label indicating whenever a  transaction is 
	identified as fraud. This label was created internally in the card processing company, and can 
	be regarded as highly accurate. 

	\begin{table}[b]
		  \centering
	    \caption{Credit card fraud detection cost matrix \cite{CorreaBahnsen2013}}
	    \label{tab:f_mat}
      \begin{tabular}{c|c|c}
				\multicolumn{1}{c|}{}  & Actual Positive& Actual Negative \\
				\multicolumn{1}{c|}{} & $y_i=1$& $y_i=0$ \\
				\hline
				Predicted Positive 		& \multirow{ 2}{*}{$C_{TP_i}=C_a$} & \multirow{ 2}{*}{$C_{FP_i}=C_a$} 
				\\
				$c_i=1$ & &\\
				\hline
				Predicted Negative  	& \multirow{ 2}{*}{$C_{FN_i}=Amt_i$} & \multirow{ 
				2}{*}{$C_{TN_i}=0$} \\
				$c_i=0$ & &\\
				%\hline
			\end{tabular}
	\end{table}
	  
\subsection{Churn modeling}

  Customer churn predictive modeling deals with estimating the probability of a customer 
	defecting using historical, behavioral and socio-economical information. The problem  of churn 
	predictive modeling has been widely studied by the data mining and machine learning communities. 
	It is usually tackled by using classification algorithms in order to learn the different patterns 
	of both the churners and non-churners. For a review see \cite{Ngai2009}. Nevertheless, current 
	state-of-the-art classification algorithms are not well aligned with commercial goals, in the 
	sense that, the models miss to include the real financial costs and benefits during the training 
	and evaluation phases \cite{Verbraken2013}.

	We then follow the example-dependent cost-sensitive methodology for churn modeling we proposed 
	in \cite{CorreaBahnsen2015a}. When a customer is predicted to be a churner, an offer is made with 
	the objective of avoiding the customer defecting. However, if a customer is actually a churner, he 
	may or not accept the offer with a probability $\gamma_i$. If the customer accepts the offer, the 
	financial impact is equal to the cost of the offer ($C_{o_i}$) plus the administrative cost of 
	contacting the customer ($C_a$). On the other hand, if the customer declines the offer, the cost 
	is the expected 	income that the clients would otherwise generate, also called customer lifetime 
	value ($CLV_i$), 	plus $C_a$. Lastly, if the customer is not actually a churner, he will be happy 
	to accept the 	offer and the cost will be $C_{o_i}$ plus $C_a$.
	In the case that the customer is predicted as non-churner, there are two possible outcomes. 
	Either the customer is not a churner, then the cost is zero, or the customer is a churner and the 
	cost is $CLV_i$. In \tablename{ \ref{tab:churn_mat}}, the cost matrix is shown.

  \begin{table}[t]
	  \centering
    \caption{Churn modeling cost matrix \cite{CorreaBahnsen2015a}}
    \label{tab:churn_mat}
    \begin{tabular}{c|c|c}
			\multicolumn{1}{c|}{}  & Actual Positive& Actual Negative \\
			\multicolumn{1}{c|}{} & $y_i=1$& $y_i=0$ \\
			\hline
			Predicted Pos 		& $C_{TP_i}=\gamma_iC_{o_i}+$ & 
													\multirow{2}{*}{$C_{FP_i}=C_{o_i}+C_a$}\\
			$c_i=1$ &$(1-\gamma_i)(CLV_i+C_a)$ &\\
			\hline
			Predicted Neg  	& \multirow{ 2}{*}{$C_{FN_i}=CLV_i$} & \multirow{ 
			2}{*}{$C_{TN_i}=0$} \\
			$c_i=0$ & &\\
			%\hline
		\end{tabular}
  \end{table}

  For this paper we used a dataset provided by a TV cable provider. The dataset consists of active 
	customers during the first semester of 2014. 	The total dataset contains 9,410 individual 
	registries, each one with 45 attributes, including a churn label indicating whenever a customer 
	is a churner.
  
\subsection{Credit scoring}
 
	The objective in credit scoring is to classify which potential customers are likely to default a 
  contracted financial obligation based on the customer's past financial experience, and with that 
  information decide whether to approve or decline a loan~\cite{He2010}. When constructing credit 
	scores, it is a common practice to use standard cost-insensitive binary classification algorithms 
	such as logistic regression, neural networks, discriminant analysis, genetic programing, decision 
	tree, among others \cite{Thomas2005}. However, in practice, the cost associated with approving a 
	bad customer is quite different from the cost associated with declining a good customer. 
	Furthermore, the costs are not constant among customers, as customers have different 
	credit line amounts, terms, and even interest rates. 

	In this paper, we used the credit scoring example-dependent cost-sensitive cost matrix we 
	proposed 	in \cite{CorreaBahnsen2014b}. The cost matrix is shown in \tablename{ \ref{tab:c_mat}}. 
	First, 	the costs of a correct classification are zero for every customer. Then, the cost of a 
	false 	negative is defined as the credit line $Cl_i$ times the loss given default $L_{gd}$. 	
	On the other hand, in the case of a false positive, the cost is the sum of $r_i$ and 	
	$C^a_{FP}$, 	where $r_i$ is the loss in profit by rejecting what would have been a good customer. 
	The second 	term $C^a_{FP}$, is related to the assumption that the financial institution will not 
	keep the 	money of the declined customer idle. It will instead give a loan to an alternative 	
	customer, and 	it is calculated as \mbox{$C^a_{FP}=- \overline{r} \cdot \pi_0+\overline{Cl}\cdot 
	L_{gd} \cdot 	\pi_1$}. 
 
  For this paper we use two different publicly available credit scoring datasets. The first    
	dataset is the \textbf{2011 Kaggle competition Give Me Some Credit}\footnote{     
	http://www.kaggle.com/c/GiveMeSomeCredit/}, in which the objective is to identify those customers 
	of personal loans that will experience financial distress in the next two years. The second 
	dataset is from the \textbf{2009 Pacific-Asia Knowledge Discovery and Data Mining conference 
	(PAKDD) competition}\footnote{http://sede.neurotech.com.br:443/PAKDD2009/}. Similarly, this 
	competition had the objective of identifying which credit card applicants were likely to default 
	and by doing so deciding whether or not to approve their applications.

  The Kaggle Credit dataset contains 112,915 examples, each one with 10 features and the class 
  label. The proportion of default or positive examples is 6.74\%. On the other hand, the PAKDD 
	Credit dataset contains 38,969 examples, with 30 features and the class label, with a proportion 
	of 19.88\% positives. This database comes from a Brazilian financial institution, and as it 
	can be inferred from the competition description, the data was obtained around 2004.

	\begin{table}[t]
		  \centering
	    \caption{Credit scoring cost matrix \cite{CorreaBahnsen2014b}}
	    \label{tab:c_mat}
      \begin{tabular}{c|c|c}
				\multicolumn{1}{c|}{}  & Actual Positive& Actual Negative \\
				\multicolumn{1}{c|}{} & $y_i=1$& $y_i=0$ \\
				\hline
				Predicted Positive 		& \multirow{ 2}{*}{$C_{TP_i}=0$} & \multirow{ 
				2}{*}{$C_{FP_i}=r_i+C^a_{FP}$} 
				\\
				$c_i=1$ & &\\
				\hline
				Predicted Negative  	& \multirow{ 2}{*}{$C_{FN_i}=Cl_i \cdot L_{gd}$} & \multirow{ 
				2}{*}{$C_{TN_i}=0$} \\
				$c_i=0$ & &\\
				%\hline
			\end{tabular}
  \end{table}
  
\subsection{Direct Marketing}
  
	In direct marketing the objective is to classify those customers who are more likely to have a 
	certain response to a marketing campaign \cite{Ngai2009}. We used a direct marketing dataset 
	from \cite{Moro2011}. The dataset contains 45,000 clients of a Portuguese bank who were 
	contacted by phone between March 2008 and October 2010 and received an offer to open a long-term 
	deposit account with attractive interest rates. The dataset contains features such as age, job, 
	marital status, education, average yearly balance and current loan status and the label 
	indicating whether or not the client accepted the offer.

	This problem is example-dependent cost sensitive, since there are different costs of false 
	positives and false negatives. Specifically, in direct marketing, false positives have the cost 
	of contacting the client, and false negatives have the cost due to the loss of income by failing 
	to contact a client that otherwise would have opened a long-term deposit. 
  
	\begin{table}[t]
		  \centering
	    \caption{Direct marketing cost matrix \cite{CorreaBahnsen2014}}
	    \label{tab:d_mat}
      \begin{tabular}{c|c|c}
				\multicolumn{1}{c|}{}  & Actual Positive& Actual Negative \\
				\multicolumn{1}{c|}{} & $y_i=1$& $y_i=0$ \\
				\hline
				Predicted Positive 		& \multirow{ 2}{*}{$C_{TP_i}=C_a$} & \multirow{ 2}{*}{$C_{FP_i}=C_a$} 
				\\
				$c_i=1$ & &\\
				\hline
				Predicted Negative  	& \multirow{ 2}{*}{$C_{FN_i}=Int_i$} & \multirow{ 2}{*}{$C_{TN_i}=0$} 
				\\
				$c_i=0$ & &\\
				%\hline
			\end{tabular}
	\end{table}
		
	We used the direct marketing example-dependent cost matrix we proposed in 
	\cite{CorreaBahnsen2014}. 
	The cost matrix is shown in \mbox{\tablename{ \ref{tab:d_mat}}}, where $C_a$ is the 
	administrative 	cost of contacting the client, and $Int_i$ is the expected income 
	when a client opens a long-term deposit. This last term is defined as the long-term deposit 
	amount times the interest rate spread.
  
\subsection{Database partitioning}  

	For each database, 3 different datasets are extracted: training, validation and testing. Each one 
	containing 50\%, 25\% and 25\% of the transactions, respectively. Afterwards, because 
	classification algorithms suffer when the label distribution is skewed towards one of the 
	classes \cite{Hastie2009},  an under-sampling of the positive	examples is made, in order to have 
  a balanced class distribution. Additionally, we perform the cost-proportionate 
  rejection-sampling and cost-proportionate over-sampling procedures. \mbox{\tablename{ 
  \ref{tab:datasets}}}, summarizes the different datasets. It is important to note that the 
  sampling procedures were only applied to the training dataset since the validation and test 
  datasets must reflect the real 	distribution.
	
	\section{Results}

	For the experiments we first used three classification algorithms, decision tree ($DT$), logistic 
  regression ($LR$) and random forest ($RF$). Using the implementation of Scikit-learn 
  \cite{Pedregosa2011}, each algorithm is trained using the different training sets: training 
  ($t$), under-sampling ($u$), cost-proportionate rejection-sampling  ($r$) \cite{Zadrozny2003}   
  and   cost-proportionate over-sampling ($o$) \cite{Elkan2001}. Afterwards,  we evaluate the 
  results of  the algorithms using $BMR$ \cite{CorreaBahnsen2014}. Then, the cost-sensitive 
  logistic  regression ($CSLR$) \cite{CorreaBahnsen2014b} and cost-sensitive decision tree 
  ($CSDT$) \cite{CorreaBahnsen2015} were also evaluated. Lastly, we calculate the 
  proposed ensembles of cost-sensitive decision trees algorithms. In particular, using each of the 
  random inducer methods, bagging ($CSB$), pasting ($CSP$), random forests ($CSRF$) and random 
  patches ($CSRP$), and then blending the base classifiers using each one of the combination 
  methods; majority voting ($mv$), cost-sensitive weighted voting ($wv$) and cost-sensitive 
  stacking ($s$). Unless otherwise stated, the random selection of the training set was repeated 50 
  times, and in each time the models were trained and results collected, this allows us to measure 
  the stability of the results. 
	
	\begin{table}[t]
    \caption{Summary of the datasets}
    \centering
    \label{tab:datasets}
		\begin{tabular}{l l c c c } %sum 7.7
		  \hline
			\textbf{Database}& \textbf{Set}&	\textbf{\# Obs} &	\textbf{\%Pos} & 
			\textbf{Cost} \\
		  \hline
		  Fraud &total&236,735&1.50&895,154\\
		  Detection&  $t$&94,599&1.51&358,078\\
		  &$u$&2,828&50.42&358,078\\
		  &$r$&94,522&1.43&357,927\\
		  &$o$&189,115&1.46&716,006\\
		  &val&70,910&1.53&274,910\\
		  &test&71,226&1.45&262,167\\
		 \hline
			Churn&total&9,410&4.83&580,884\\
			Modeling&$t$&3,758&5.05&244,542\\
			&$u$ &374&50.80&244,542\\
			&$r$&428&41.35&431,428\\
			&$o$ &5,767&31.24&2,350,285\\
			&val&2,824&4.77&174,171\\
			&test&2,825&4.42&162,171\\
			\hline
		  Credit  & total&112,915&6.74&83,740,181\\
		  Scoring1 & $t$&45,264&6.75&33,360,130\\
		  &$u$&6,038&50.58&33,360,130\\
		  &$r$&5,271&43.81&29,009,564\\
		  &$o$&66,123&36.16&296,515,655\\
		  &val&33,919&6.68&24,786,997\\
		  &test&33,732&6.81&25,593,055\\
		  \hline
		  Credit &total&38,969&19.88&3,117,960\\
			Scoring 2&$t$&15,353&19.97&1,221,174\\
			&$u$&6,188&49.56&1,221,174\\
			&$r$&2,776&35.77&631,595\\
			&$o$&33,805&33.93&6,798,282\\
			&val&11,833&20.36&991,795\\
			&test&11,783&19.30&904,991\\
		  \hline
		  Direct &total&37,931&12.62&59,507\\
	    Marketing&$t$&15,346&12.55&24,304\\
	    &$u$&3,806&50.60&24,304\\
	    &$r$&1,644&52.43&20,621\\
	    &$o$&22,625&40.69&207,978\\
	    &val&11,354&12.30&16,154\\
	    &test&11,231&13.04&19,048\\
		  \hline
	  \end{tabular}
  \end{table}
 
	The results are shown in \tablename{ \ref{tab:results_savings}}. First, when observing 	the 
	results of the cost-insensitive methods ($CI$), that is, $DT$, $LR$ and $RF$ algorithms trained 
	on the $t$ and $u$ sets, the $RF$ algorithm produces the best result by savings in three out of 
	the five sets, followed by the $LR-u$. It is also clear that the results on the $t$ dataset are 
	not as good as the ones on the $u$, this is highly related to the unbalanced distribution of the 
	positives and negatives in all the databases.

	In the case of cost-proportionate sampling methods ($CPS$), specifically the 
	cost-proportionate rejection sampling ($r$) and cost-proportionate over 
	sampling ($o$). It is observed than in four cases the savings increases quite 
	significantly. It is on the fraud detection database where these methods do not outperform the 
	algorithms trained on the under-sampled set. This may be related to the fact that in this 
	database the initial percentage of positives is 1.5\% which is similar to the percentage in the 
	$r$ and 	$o$ sets. However it is 50.42\% in the $u$ set, which may help explain why this method 
	performs much better as measured by savings.

	Afterwards, in the case of the $BMR$ algorithms, the results show that this method outperforms 
	the previous ones in four cases and has almost the same result in the other set. In the fraud 
	detection	 set, the results are quite better, since the savings of the three classification 
	algorithms increase when using this methodology. The next family of algorithms is the 
	cost-sensitive training, which includes the $CSLR$ and $CSDT$ techniques. In this case, only in 
	two databases the results are improved. Lastly, we evaluate the proposed $ECSDT$ algorithms. The 
	results show that these methods arise to the best overall results in three sets, while being 
	quite competitive in the others.

	\begin{table*}[t]
		\caption{Results of the algorithms measured by savings}
		\centering
		\label{tab:results_savings}
		\begin{tabular}{l l r@{\hskip 0in}c@{\hskip 0in}l r@{\hskip 0in}c@{\hskip 0in}l r@{\hskip 
		0in}c@{\hskip 0in}l r@{\hskip 0in}c@{\hskip 0in}l r@{\hskip 0in}c@{\hskip 0in}l  } %sum 7.7
		\hline
		\bf{Family} & \bf{Algorithm} & \multicolumn{3}{c}{\bf{Fraud}} & 
		\multicolumn{3}{c}{\bf{Churn}} & \multicolumn{3}{c}{\bf{Credit 1}} & 
		\multicolumn{3}{c}{\bf{Credit 2}} & \multicolumn{3}{c}{\bf{Marketing}} \\ 
		\hline
CI&DT-t & 0.3176 &$\pm$& 0.0357 & -0.0018 &$\pm$& 0.0194 & 0.1931 &$\pm$& 0.0087 & -0.0616 &$\pm$& 
0.0229 & -0.2342 &$\pm$& 0.0609\\ 
&LR-t & 0.0092 &$\pm$& 0.0002 & -0.0001 &$\pm$& 0.0002 & 0.0177 &$\pm$& 0.0126 & 0.0039 &$\pm$& 
0.0012 & -0.2931 &$\pm$& 0.0602\\ 
&RF-t & 0.3342 &$\pm$& 0.0156 & -0.0026 &$\pm$& 0.0079 & 0.1471 &$\pm$& 0.0071 & 0.0303 &$\pm$& 
0.0040 & -0.2569 &$\pm$& 0.0637\\ 
&DT-u & 0.5239 &$\pm$& 0.0118 & -0.0389 &$\pm$& 0.0583 & 0.3287 &$\pm$& 0.0125 & -0.1893 &$\pm$& 
0.0314 & -0.0278 &$\pm$& 0.0475\\ 
&LR-u & 0.1243 &$\pm$& 0.0387 & 0.0039 &$\pm$& 0.0492 & 0.4118 &$\pm$& 0.0313 & 0.1850 &$\pm$& 
0.0231 & 0.2200 &$\pm$& 0.0376\\ 
&RF-u & 0.5684 &$\pm$& 0.0097 & 0.0433 &$\pm$& 0.0533 & 0.4981 &$\pm$& 0.0079 & 0.1237 &$\pm$& 
0.0228 & 0.1227 &$\pm$& 0.0443\\ 
\hline 
CPS&DT-r & 0.3439 &$\pm$& 0.0453 & 0.0054 &$\pm$& 0.0568 & 0.3310 &$\pm$& 0.0126 & 0.0724 &$\pm$& 
0.0212 & 0.1960 &$\pm$& 0.0527\\ 
&LR-r & 0.3077 &$\pm$& 0.0301 & 0.0484 &$\pm$& 0.0375 & 0.3965 &$\pm$& 0.0263 & 0.2650 &$\pm$& 
0.0115 & 0.4210 &$\pm$& 0.0267\\ 
&RF-r & 0.3812 &$\pm$& 0.0264 & 0.1056 &$\pm$& 0.0412 & 0.4989 &$\pm$& 0.0080 & 0.3055 &$\pm$& 
0.0106 & 0.3840 &$\pm$& 0.0360\\ 
&DT-o & 0.3172 &$\pm$& 0.0274 & 0.0251 &$\pm$& 0.0195 & 0.1738 &$\pm$& 0.0092 & 0.0918 &$\pm$& 
0.0225 & -0.2598 &$\pm$& 0.0559\\ 
&LR-o & 0.2793 &$\pm$& 0.0185 & 0.0316 &$\pm$& 0.0228 & 0.3301 &$\pm$& 0.0109 & 0.2554 &$\pm$& 
0.0090 & 0.3129 &$\pm$& 0.0277\\ 
&RF-o & 0.3612 &$\pm$& 0.0295 & 0.0205 &$\pm$& 0.0156 & 0.2128 &$\pm$& 0.0081 & 0.2242 &$\pm$& 
0.0070 & -0.1782 &$\pm$& 0.0618\\ 
\hline 
BMR&DT-t-BMR & 0.6045 &$\pm$& 0.0386 & 0.0298 &$\pm$& 0.0145 & 0.1054 &$\pm$& 0.0358 & 0.2740 
&$\pm$& 0.0067 & 0.4598 &$\pm$& 0.0089\\ 
&LR-t-BMR & 0.4552 &$\pm$& 0.0203 & 0.1082 &$\pm$& 0.0316 & 0.2189 &$\pm$& 0.0541 & \bf{0.3148} 
&\bf{$\pm$}& \bf{0.0094} & \bf{0.4973} &\bf{$\pm$}& \bf{0.0084}\\ 
&RF-t-BMR & 0.6414 &$\pm$& 0.0154 & 0.0856 &$\pm$& 0.0354 & 0.4924 &$\pm$& 0.0087 & 0.3133 &$\pm$& 
0.0094 & 0.4807 &$\pm$& 0.0093\\ 
\hline 
CST&CSLR-t & 0.6113 &$\pm$& 0.0262 & 0.1118 &$\pm$& 0.0484 & 0.4554 &$\pm$& 0.1039 & 0.2748 &$\pm$& 
0.0069 & 0.4484 &$\pm$& 0.0072\\ 
&CSDT-t & 0.7116 &$\pm$& 0.2557 & 0.1115 &$\pm$& 0.0378 & 0.4829 &$\pm$& 0.0098 & 0.2835 &$\pm$& 
0.0078 & 0.4741 &$\pm$& 0.0063\\ 
\hline 
ECSDT&CSB-mv-t & 0.7124 &$\pm$& 0.0162 & 0.1237 &$\pm$& 0.0368 & 0.4862 &$\pm$& 0.0102 & 0.2945 
&$\pm$& 0.0105 & 0.4837 &$\pm$& 0.0078\\ 
&CSB-wv-t & 0.7276 &$\pm$& 0.0116 & 0.1539 &$\pm$& 0.0255 & 0.4862 &$\pm$& 0.0102 & 0.2948 &$\pm$& 
0.0106 & 0.4838 &$\pm$& 0.0079\\ 
&CSB-s-t & 0.7181 &$\pm$& 0.0109 & 0.1441 &$\pm$& 0.0364 & 0.4847 &$\pm$& 0.0096 & 0.2856 &$\pm$& 
0.0088 & 0.4769 &$\pm$& 0.0078\\ 
&CSP-mv-t & 0.7106 &$\pm$& 0.0113 & 0.1227 &$\pm$& 0.0399 & 0.4853 &$\pm$& 0.0104 & 0.2919 &$\pm$& 
0.0097 & 0.4831 &$\pm$& 0.0081\\ 
&CSP-wv-t & 0.7244 &$\pm$& 0.0202 & 0.1501 &$\pm$& 0.0302 & 0.4854 &$\pm$& 0.0105 & 0.2921 &$\pm$& 
0.0098 & 0.4832 &$\pm$& 0.0082\\ 
&CSP-s-t & 0.7212 &$\pm$& 0.0067 & 0.1488 &$\pm$& 0.0272 & 0.4848 &$\pm$& 0.0084 & 0.2870 &$\pm$& 
0.0084 & 0.4752 &$\pm$& 0.0089\\ 
&CSRF-mv-t & 0.6498 &$\pm$& 0.0598 & 0.0300 &$\pm$& 0.0488 & 0.4980 &$\pm$& 0.0120 & 0.2274 &$\pm$& 
0.0520 & 0.3929 &$\pm$& 0.0655\\ 
&CSRF-wv-t & 0.7249 &$\pm$& 0.0742 & 0.0624 &$\pm$& 0.0477 & 0.4979 &$\pm$& 0.0124 & 0.2948 &$\pm$& 
0.0079 & 0.4728 &$\pm$& 0.0125\\ 
&CSRF-s-t & 0.6731 &$\pm$& 0.0931 & 0.0586 &$\pm$& 0.0507 & 0.4839 &$\pm$& 0.0160 & 0.2518 &$\pm$& 
0.0281 & 0.3854 &$\pm$& 0.0899\\ 
&CSRP-mv-t & 0.7220 &$\pm$& 0.0082 & 0.1321 &$\pm$& 0.0280 & \bf{0.5154} &\bf{$\pm$}& \bf{0.0077} & 
0.3053 &$\pm$& 0.0087 & 0.4960 &$\pm$& 0.0075\\ 
&CSRP-wv-t & \bf{0.7348} &\bf{$\pm$}& \bf{0.0131} & 0.1615 &$\pm$& 0.0252 & 0.5152 &$\pm$& 0.0083 & 
0.3015 &$\pm$& 0.0086 & 0.4885 &$\pm$& 0.0076\\ 
&CSRP-s-t & 0.7336 &$\pm$& 0.0108 & \bf{0.1652} &\bf{$\pm$}& \bf{0.0264} & 0.4989 &$\pm$& 0.0088 & 
0.2956 &$\pm$& 0.0078 & 0.4878 &$\pm$& 0.0080\\ 
	\hline
	\multicolumn{17}{c}{(those models with the highest savings are market as bold)}
	\end{tabular}
	\end{table*}

	  \begin{figure}[t]
    \centering
    \includegraphics{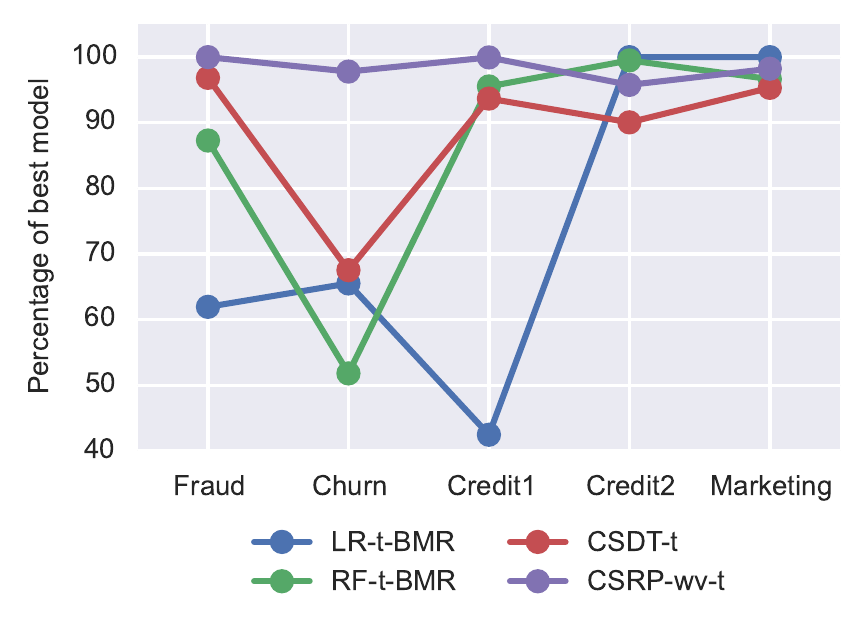}
    \caption{\textbf{Comparison of the savings of the algorithms versus the highest savings in 
    each database.} The $CSRP-wt$ is very close to the best result in all the databases. 
    Additionally, even though the $LR-BMR$ is the best algorithm in two databases, the performance 
    in the other three is very poor.}
    \label{fig_comparison_best}
  \end{figure}

	Subsequently, in order to statistically sort the classifiers we computed the Friedman ranking 
	(F-Rank)	statistic \cite{Demsar2006}. This rank increases with the cost of the algorithms. 
	We also calculate the average savings of each algorithm compared with the highest savings in 
	each set (perBest), as a 	measure of how close are the savings of an algorithm to the best 
	result. In \tablename{ \ref{tab:results_ranking}}, the results are shown. It is observed that the 
	first six algorithms, according to the F-Rank, belong to the $ECSDT$ family. In particular, the 
	best three classifiers is the ensemble of cost-sensitive decision trees using the random patches 
	approach. Giving the best result the one that blends the base classifiers using weighted voting 
	method. Moreover as shown in  \tablename{ \ref{tab:results_best}}, this method ranks on each 
	dataset \nth{1}, \nth{2}, \nth{2}, \nth{5} and \nth{3}, respectively. For comparison the best 
	method from an other family is the $RF$ with $BMR$, which ranks \nth{14}, \nth{14}, \nth{8}, 
	\nth{2} and \nth{9}.
	
	\begin{table}[!t]
    \caption{Savings Friedman ranking and average percentage of best result}
    \centering
    \label{tab:results_ranking}
		\begin{tabular}{ll c c  } %sum 7.7
		  \hline
		  \bf{Family} & \bf{Algorithm} & \bf{F-Rank} & \bf{perBest}\\
		  \hline
				ECSDT&CSRP-wv-t&2.6&98.35\\ 
				ECSDT&CSRP-s-t&3.4&97.72\\ 
				ECSDT&CSRP-mv-t&4.0&94.99\\ 
				ECSDT&CSB-wv-t&5.6&95.49\\ 
				ECSDT&CSP-wv-t&7.4&94.72\\ 
				ECSDT&CSB-mv-t&8.2&91.39\\ 
				ECSDT&CSRF-wv-t&9.4&84.35\\ 
				BMR&RF-t-BMR&9.4&86.16\\ 
				ECSDT&CSP-s-t&9.6&93.80\\ 
				ECSDT&CSP-mv-t&10.2&91.00\\ 
				ECSDT&CSB-s-t&10.2&93.12\\ 
				BMR&LR-t-BMR&11.2&73.98\\ 
				CPS&RF-r&11.6&77.37\\ 
				CST&CSDT-t&12.6&88.69\\ 
				CST&CSLR-t&14.4&83.34\\ 
				ECSDT&CSRF-mv-t&15.2&70.88\\ 
				ECSDT&CSRF-s-t&16.0&75.68\\ 
				CI&RF-u&17.2&52.83\\ 
				CPS&LR-r&19.0&63.39\\ 
				BMR&DT-t-BMR&19.0&60.05\\ 
				CPS&LR-o&21.0&53.05\\ 
				CPS&DT-r&22.6&35.33\\ 
				CI&LR-u&22.8&40.43\\ 
				CPS&RF-o&22.8&34.81\\ 
				CI&DT-u&24.4&27.01\\ 
				CPS&DT-o&25.0&24.25\\ 
				CI&DT-t&26.0&16.14\\ 
				CI&RF-t&26.2&16.73\\ 
				CI&LR-t&28.0&1.19\\ 
		  \hline
	  \end{tabular}
  \end{table}

 \begin{figure*}[t]
	\centering
	\subfloat[Comparison of the Friedman ranking.]{\includegraphics{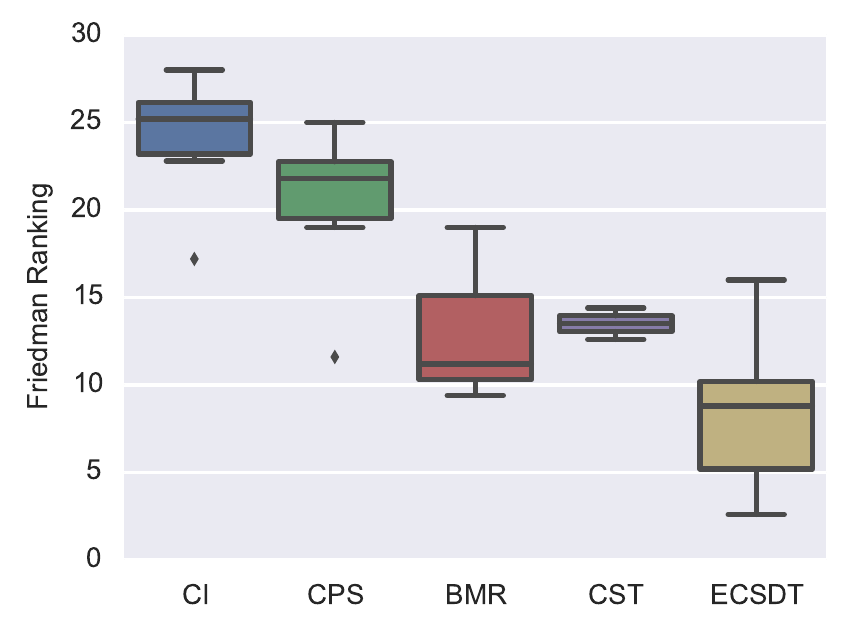}%
		\label{fig_1}}
	\hfil
	\subfloat[Comparison of the average savings of the algorithms versus the 
		highest savings.]{\includegraphics{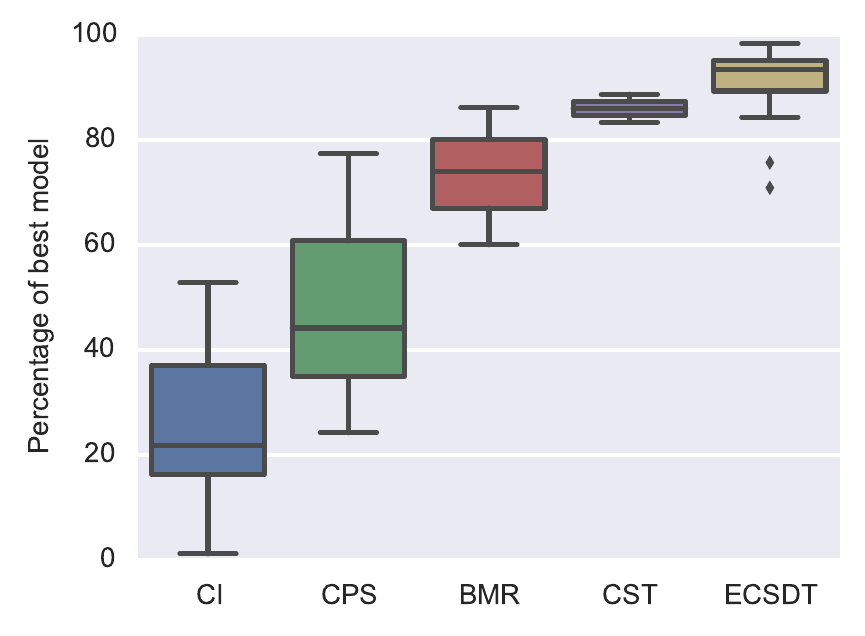}%
		\label{fig_comparison_best_by_family}}
	\caption{\textbf{Comparison of the results by family of classifiers.} The $ECSDT$ family has the 
	best performance measured either by Friedman ranking or average percentage of best model.}
	\label{fig_comparison_family}
\end{figure*}

	\begin{table}[t]
    \caption{Savings ranks of best algorithm of each family by database}
    \centering
    \label{tab:results_best}
		\begin{tabular}{l c c c c c  } %sum 7.7
		  \hline
		  \bf{Algorithm} & \bf{Fraud} & \bf{Churn} &\bf{Credit1} & \bf{Credit2} & \bf{Marketing} \\
		  \hline
			RF-u&17&18&5&23&23\\ 
			RF-r&20&13&3&3&19\\ 
			RF-t-BMR&14&14&8&2&9\\ 
			CSDT-t&10&11&16&14&12\\ 
			CSRP-wv-t&1&2&2&5&3\\ 
		  \hline
	  \end{tabular}
  \end{table}
  
 \begin{figure*}[!t]
	\centering
	\subfloat[Comparison by inducer methodology.]{\includegraphics{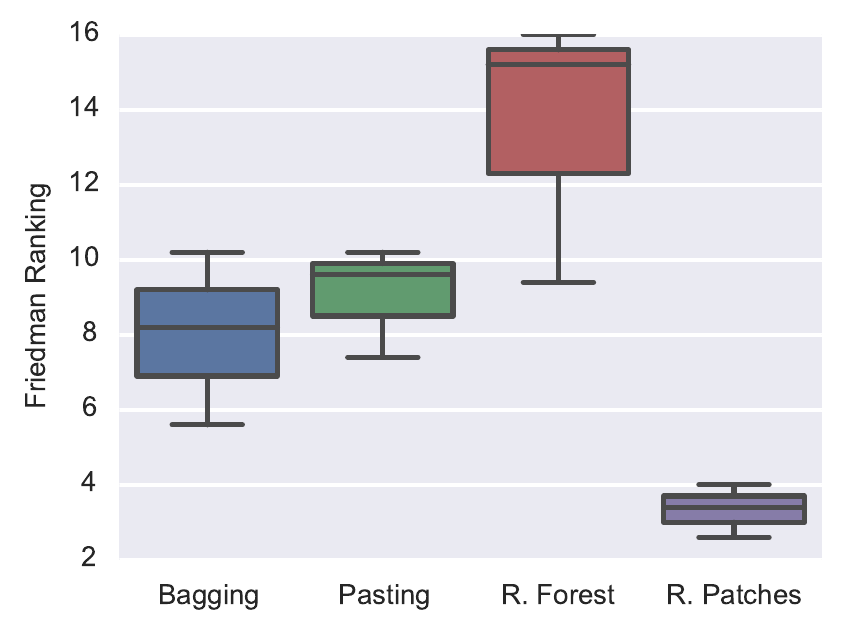}%
		\label{fig_rank_ecsdt1}}
	\hfil
	\subfloat[Comparison by combination of base classifiers approach.]{\includegraphics{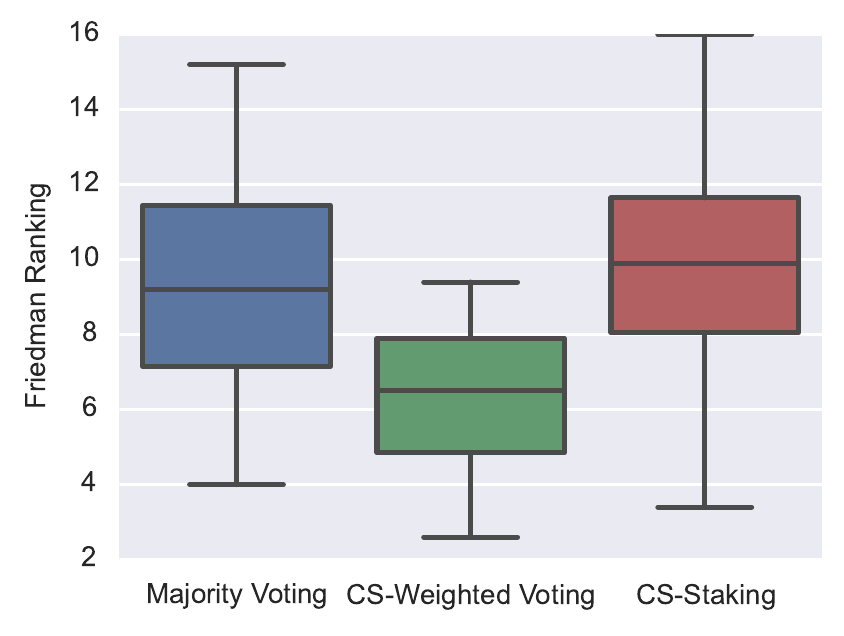}%
		\label{fig_rank_ecsdt2}}
	\caption{\textbf{Comparison of the Friedman ranking within the $ECSDT$ family.} Overall, the 
	random inducer method that provides the best results is the $CSRP$. Moreover, the best 
	combination method compared by Friedman ranking is the cost-sensitive weighted voting.}
	\label{fig_rank_ecsdt}
\end{figure*}

    Moreover, when analyzing the perBest statistic, it is observed that it follows almost the same 
  order as the F-Rank. Notwithstanding, there are cases in which algorithms ranks are different in 
  the two statistics, for example the $CSDT-t$ algorithm has a lower F-Rank than the $RF-BMR$, 
  but the perBest if better. This happens because, the F-Rank does not take into account the 
  difference in savings within algorithms. This can be further investigated in \figurename{ 
  \ref{fig_comparison_best}}. Even that ranks of the $BMR$ models are better than the $CSDT$, the 
  latter is on average closer to the best performance method in each set. Moreover, it is confirmed 
  that the $CSRP-wt$ is very close to the best result in all cases. Lastly, it is shown why the 
  F-Rank of the $LR-BMR$ is high, given the fact that is the best model in two databases. The 
  reason  for that, is because the performance on the other sets is very poor.
  
% \begin{figure}[!t]
%   \centering
%   \includegraphics{fig_4}
%   \caption{Comparison of the Friedman ranking within the $ECSDT$ family of algorithms by inducer 
% 	methodology.}
%   \label{fig_rank_ecsdt1}
%   \end{figure} 
%   
% \begin{figure}[!t]
%   \centering
%   \includegraphics{fig_3}
%   \caption{Comparison of the Friedman ranking within the $ECSDT$ family of algorithms by 
% combination 
% of base classifiers approach.}
%   \label{fig_rank_ecsdt2}
%   \end{figure} 

\begin{table*}[t]
    \caption{Results as measured by F1Score}
    \centering
    \label{tab:results_f1score}
		\begin{tabular}{l l r@{\hskip 0in}c@{\hskip 0in}l r@{\hskip 0in}c@{\hskip 0in}l r@{\hskip 
0in}c@{\hskip 0in}l r@{\hskip 0in}c@{\hskip 0in}l r@{\hskip 0in}c@{\hskip 0in}l  } %sum 7.7
		  \hline
		\bf{Family} & \bf{Algorithm} & \multicolumn{3}{c}{\bf{Fraud}} & 
\multicolumn{3}{c}{\bf{Churn}} & \multicolumn{3}{c}{\bf{Credit 1}} & \multicolumn{3}{c}{\bf{Credit 
2}} & \multicolumn{3}{c}{\bf{Marketing}} \\ 
		  \hline
CI&DT-t & 0.4458 &$\pm$& 0.0133 & 0.0733 &$\pm$& 0.0198 & 0.2593 &$\pm$& 0.0068 & 0.2614 &$\pm$& 
0.0083 & 0.2647 &$\pm$& 0.0079\\ 
&LR-t & 0.1531 &$\pm$& 0.0045 & 0.0000 &$\pm$& 0.0000 & 0.0494 &$\pm$& 0.0277 & 0.0155 &$\pm$& 
0.0037 & 0.2702 &$\pm$& 0.0125\\ 
&RF-t & 0.2061 &$\pm$& 0.0041 & 0.0249 &$\pm$& 0.0146 & 0.2668 &$\pm$& 0.0085 & 0.0887 &$\pm$& 
0.0061 & 0.2884 &$\pm$& 0.0116\\ 
&DT-u & 0.1502 &$\pm$& 0.0066 & 0.1175 &$\pm$& 0.0103 & 0.2276 &$\pm$& 0.0044 & 0.3235 &$\pm$& 
0.0055 & 0.2659 &$\pm$& 0.0061\\ 
&LR-u & 0.0241 &$\pm$& 0.0163 & 0.1222 &$\pm$& 0.0098 & 0.3160 &$\pm$& 0.0314 & \bf{0.3890} 
&\bf{$\pm$}& \bf{0.0053} & 0.3440 &$\pm$& 0.0083\\ 
&RF-u & 0.0359 &$\pm$& 0.0065 & 0.1346 &$\pm$& 0.0112 & 0.3193 &$\pm$& 0.0053 & 0.3815 &$\pm$& 
0.0051 & 0.3088 &$\pm$& 0.0065\\ 
\hline 
CPS&DT-r & 0.4321 &$\pm$& 0.0086 & 0.1206 &$\pm$& 0.0132 & 0.2310 &$\pm$& 0.0049 & 0.3409 &$\pm$& 
0.0046 & 0.2739 &$\pm$& 0.0076\\ 
&LR-r & 0.1846 &$\pm$& 0.0123 & 0.1258 &$\pm$& 0.0111 & 0.3597 &$\pm$& 0.0156 & 0.3793 &$\pm$& 
0.0049 & 0.3374 &$\pm$& 0.0101\\ 
&RF-r & 0.2171 &$\pm$& 0.0100 & 0.1450 &$\pm$& 0.0131 & 0.3361 &$\pm$& 0.0067 & 0.3570 &$\pm$& 
0.0048 & 0.3103 &$\pm$& 0.0075\\ 
&DT-o & \bf{0.4495} &\bf{$\pm$}& \bf{0.0063} & 0.1022 &$\pm$& 0.0180 & 0.2459 &$\pm$& 0.0081 & 
0.3258 &$\pm$& 0.0056 & 0.2634 &$\pm$& 0.0089\\ 
&LR-o & 0.1776 &$\pm$& 0.0117 & 0.1085 &$\pm$& 0.0203 & 0.3769 &$\pm$& 0.0067 & 0.3804 &$\pm$& 
0.0044 & \bf{0.3568} &\bf{$\pm$}& \bf{0.0102}\\ 
&RF-o & 0.2129 &$\pm$& 0.0080 & 0.0841 &$\pm$& 0.0201 & 0.3281 &$\pm$& 0.0078 & 0.3212 &$\pm$& 
0.0054 & 0.3083 &$\pm$& 0.0093\\ 
\hline 
BMR&DT-t-BMR & 0.2139 &$\pm$& 0.0215 & 0.0941 &$\pm$& 0.0157 & 0.1514 &$\pm$& 0.0390 & 0.3338 
&$\pm$& 0.0052 & 0.2433 &$\pm$& 0.0071\\ 
&LR-t-BMR & 0.1384 &$\pm$& 0.0044 & 0.1370 &$\pm$& 0.0150 & 0.1915 &$\pm$& 0.0340 & 0.3572 &$\pm$& 
0.0045 & 0.2954 &$\pm$& 0.0079\\ 
&RF-t-BMR & 0.2052 &$\pm$& 0.0183 & 0.1264 &$\pm$& 0.0156 & 0.3186 &$\pm$& 0.0072 & 0.3551 &$\pm$& 
0.0053 & 0.2744 &$\pm$& 0.0070\\ 
\hline 
CST&CSLR-t & 0.2031 &$\pm$& 0.0065 & 0.1134 &$\pm$& 0.0151 & 0.1454 &$\pm$& 0.0517 & 0.3363 &$\pm$& 
0.0045 & 0.2339 &$\pm$& 0.0051\\ 
&CSDT-t & 0.2522 &$\pm$& 0.0980 & 0.1288 &$\pm$& 0.0194 & 0.2754 &$\pm$& 0.0059 & 0.3483 &$\pm$& 
0.0046 & 0.2680 &$\pm$& 0.0060\\ 
\hline 
ECSDT&CSB-mv-t & 0.2112 &$\pm$& 0.0125 & 0.1481 &$\pm$& 0.0122 & 0.2927 &$\pm$& 0.0108 & 0.3503 
&$\pm$& 0.0046 & 0.2758 &$\pm$& 0.0072\\ 
&CSB-wv-t & 0.2112 &$\pm$& 0.0091 & 0.1686 &$\pm$& 0.0125 & 0.2926 &$\pm$& 0.0108 & 0.3503 &$\pm$& 
0.0046 & 0.2757 &$\pm$& 0.0072\\ 
&CSB-s-t & 0.2072 &$\pm$& 0.0103 & 0.1554 &$\pm$& 0.0121 & 0.2818 &$\pm$& 0.0075 & 0.3533 &$\pm$& 
0.0046 & 0.2799 &$\pm$& 0.0074\\ 
&CSP-mv-t & 0.2098 &$\pm$& 0.0126 & 0.1480 &$\pm$& 0.0136 & 0.2903 &$\pm$& 0.0108 & 0.3499 &$\pm$& 
0.0044 & 0.2749 &$\pm$& 0.0064\\ 
&CSP-wv-t & 0.2099 &$\pm$& 0.0054 & 0.1651 &$\pm$& 0.0120 & 0.2905 &$\pm$& 0.0110 & 0.3498 &$\pm$& 
0.0045 & 0.2749 &$\pm$& 0.0064\\ 
&CSP-s-t & 0.2064 &$\pm$& 0.0069 & 0.1590 &$\pm$& 0.0099 & 0.2809 &$\pm$& 0.0062 & 0.3524 &$\pm$& 
0.0046 & 0.2778 &$\pm$& 0.0104\\ 
&CSRF-mv-t & 0.2208 &$\pm$& 0.0022 & 0.1081 &$\pm$& 0.0224 & 0.2994 &$\pm$& 0.0226 & 0.3560 &$\pm$& 
0.0118 & 0.2780 &$\pm$& 0.0106\\ 
&CSRF-wv-t & 0.2175 &$\pm$& 0.0019 & 0.1220 &$\pm$& 0.0234 & 0.2992 &$\pm$& 0.0236 & 0.3549 &$\pm$& 
0.0069 & 0.2552 &$\pm$& 0.0187\\ 
&CSRF-s-t & 0.2169 &$\pm$& 0.0045 & 0.1304 &$\pm$& 0.0162 & 0.2916 &$\pm$& 0.0236 & 0.3540 &$\pm$& 
0.0071 & 0.2578 &$\pm$& 0.0186\\ 
&CSRP-mv-t & 0.2691 &$\pm$& 0.0054 & 0.1511 &$\pm$& 0.0110 & 0.4031 &$\pm$& 0.0079 & 0.3743 &$\pm$& 
0.0050 & 0.2916 &$\pm$& 0.0062\\ 
&CSRP-wv-t & 0.2780 &$\pm$& 0.0041 & \bf{0.1710} &\bf{$\pm$}& \bf{0.0140} & \bf{0.4049} 
&\bf{$\pm$}& 
\bf{0.0066} & 0.3721 &$\pm$& 0.0051 & 0.2781 &$\pm$& 0.0063\\ 
&CSRP-s-t & 0.2735 &$\pm$& 0.0148 & 0.1622 &$\pm$& 0.0103 & 0.3953 &$\pm$& 0.0141 & 0.3720 &$\pm$& 
0.0049 & 0.2922 &$\pm$& 0.0137\\ 
		  \hline
      \multicolumn{17}{c}{(those models with the highest F1Score are market as bold)}
	  \end{tabular}
  \end{table*}
  
  Furthermore \figurename{ \ref{fig_1}}, shows the Friedman ranking of each family of classifiers. 
  The $ECSDT$ methods are overall better, followed by the $BMR$ and the $CST$ methods. As 
  expected, the $CI$ family is the one that performs the worst. Nevertheless, there is 
  a significant variance within the ranks in the $ECSDT$ family, as the best one has a Friedman 
  ranking of 2.6 and the worst 16. Similar results are found when observing the perBest shown in 
  \figurename{  \ref{fig_comparison_best_by_family}}. However, in the case of the perBest, the 
  $CST$ methods perform better than the $BMR$. It is important, in both cases it is confirmed 
  that the $ECSDT$ family of methods is the one that arise to the best results as measured by 
  savings.

  We further investigate the different methods that compose the $ECSDT$ family, first by inducer 
  methods and by the combination approach. In \figurename{ \ref{fig_rank_ecsdt1}}, the Friedman 
  ranking of the $ECSDT$ methods grouped by inducer algorithm are shown. It is observed that the 
  worst method is the random forest methodology. This may be related to the fact that within the 
  random inducer methods, this is the only one that also modified the learner algorithm in the 
  sense that it randomly select features for each step during the decision tree growing. Moreover, 
  as expected the bagging and pasting methods perform quite similar, after all the only difference 
  is that in bagging the sampling is done with replacement, while it is not the case in pasting. 
  In  general the best methodology is random patches. Additionally, in \figurename{ 
  \ref{fig_rank_ecsdt2}}, a similar analysis is made taking into account the combination of base 
  classifiers approach. In this case, the best combination method is weighted voting, while 
  majority voting and staking have a similar performance.

    \begin{figure}[t]
  \centering
  \includegraphics{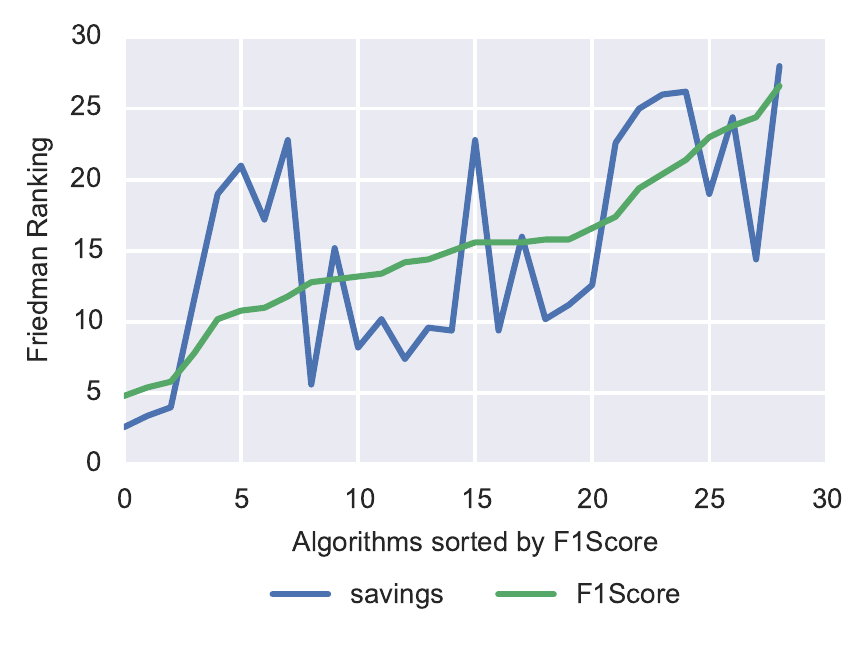}
  \caption{\textbf{Comparison of the Friedman ranking of the savings and F1Score sorted by F1Score 
  ranking.} The best two algorithms according to their Friedman rank of F1Score are indeed the 
  best ones measured by the Friedman rank of the savings. However, this relation does not 
  consistently hold for the other algorithms as the correlation between the rankings is just 
  65.10\%.}
  \label{fig_2}
  \end{figure}
  
	Finally, in \tablename{ \ref{tab:results_f1score}} the results of the algorithms measured by 
	F1Score are shown. It is observed that the model with the highest savings is not the same as the 
	one with the highest F1Score in all of the databases, corroborating the conclusions from 
	\cite{CorreaBahnsen2013}, as selecting a method by a traditional statistic does not give the 
	same result as selecting it using a business oriented measure such as financial savings. This can 
	be 	further examined in \mbox{\figurename{ \ref{fig_2}}}, where the ranking of the F1Score and 
	savings are compared. It is observed that the best two algorithms according to their Friedman 
	rank of F1Score are indeed the best ones measured by the Friedman rank of the savings. However, 
	this relation does not consistently hold for the other algorithms as the correlation between 
	the rankings is just 65.10\%.
	
\newpage
	
\section{Conclusions and future work}

	In this paper we proposed a new framework of ensembles of example-dependent cost-sensitive 
  decision-trees by creating cost-sensitive decision trees using four different 
  random inducer methods and then blending them using three different combination approaches.
	The proposed method was tested using five databases, from four real-world applications: credit 
  card fraud detection, churn modeling, credit scoring and direct marketing. We have shown 
  theoretically and experimentally that our method ranks the best and outperforms   
  state-of-the-art example-dependent cost-sensitive methodologies, when measured by financial 
  savings.
	
  In total, our framework is composed of 12 different algorithms, since the example-dependent 
  cost-sensitive ensemble can be constructed by inducing the base classifiers using either 
  bagging, pasting, random forest or random patches, and then blending them using majority voting, 
  cost-sensitive weighted voting or cost-sensitive stacking. When analyzing the results within our 
  proposed framework, it is observed that the inducer method that performs the best is random 
  patches algorithm. Furthermore, the random patches algorithm is the one with the 
  lowest complexity as each base classifier is learned on a smaller subset than with the 
  other inducer methods. Nevertheless, there is no clear winner among the different combination  
  methods. Since the most time consuming step is inducing and constructing the base classifiers, 
  testing all combination methods does not add a significant additional complexity.
	
	Our results show the importance of using the real example-dependent financial costs 
	associated with real-world applications. In particular, we found significant differences in the 
  results when evaluating a model using a  traditional cost-insensitive measure such as the 
	accuracy or F1Score,  than when using the savings. The final conclusion is that it is important 
  to use the real practical financial costs of   each context.
  
  To improve the results, future research should be focused on developing an 
example-dependent cost-sensitive boosting approach. For some applications boosting methods have 
proved to outperform the bagging algorithms. Moreover, the methods covered in this work are all 
batch, in the sense that the batch algorithms keeps the system weights constant while calculating 
the evaluation measures. However in some applications such as fraud detection, the 
evolving patters due to change in the fraudsters behavior is not capture by using batch methods. 
Therefore, the need for investigate this problem from an online-learning perspective.
	
\section*{Acknowledgments}
  Funding for this research was provided by the Fonds National de la Recherche, Luxembourg.

\bibliographystyle{myIEEEtran}

\end{document}